\documentclass[10pt,twocolumn,letterpaper]{article}

\usepackage{iccv}
\usepackage{times}
\usepackage{epsfig}
\usepackage{graphicx}
\usepackage{amsmath}
\usepackage{amssymb}

\usepackage{mathrsfs}
\usepackage{multirow}
\usepackage{algorithm}
\usepackage{algpseudocode}
\usepackage{authblk}

\usepackage[pagebackref=true,breaklinks=true,letterpaper=true,colorlinks,bookmarks=false]{hyperref}

\iccvfinalcopy 

\def\iccvPaperID{2} 

\ificcvfinal\fi
\begin{document}

\title{A Generalized and Robust Method Towards Practical Gaze Estimation on Smart Phone}

\makeatletter
\newcommand{\printfnsymbol}[1]{%
  \textsuperscript{\@fnsymbol{#1}}%
}
\makeatother

\author{Tianchu Guo\textsuperscript{1}\thanks{equal contribution}, Yongchao Liu\textsuperscript{1}\printfnsymbol{1}, Hui Zhang\textsuperscript{1}, Xiabing Liu\textsuperscript{1}, Youngjun Kwak\textsuperscript{2}, Byung In Yoo\textsuperscript{2}, Jae-Joon Han\textsuperscript{2}, Changkyu Choi\textsuperscript{2}\\
\textsuperscript{1}{Samsung Research China - Beijing} \textsuperscript{2}{Samsung Advanced Institute of Technology}\\
{\tt\ \{tianc.guo, yongchao.liu, hui123.zhang, xiabing.liu, yjk.kwak, byungin.yoo, jae-joon.han, changkyu\_choi\}@samsung.com}
}


\maketitle
\ificcvfinal\thispagestyle{empty}\fi

\begin{abstract}
   Gaze estimation for ordinary smart phone, \eg estimating where the user is looking at on the phone screen, can be applied in various applications. However, the widely used appearance-based CNN methods still have two issues for practical adoption.
   First, due to the limited dataset, gaze estimation is very likely to suffer from over-fitting, leading to poor accuracy at run time.
   Second, the current methods are usually not robust, \ie their prediction results having notable jitters even when the user is performing gaze fixation, which degrades user experience greatly.
   For the first issue, we propose a new tolerant and talented (TAT) training scheme, which is an iterative random knowledge distillation framework enhanced with cosine similarity pruning and aligned orthogonal initialization. The knowledge distillation is a tolerant teaching process providing diverse and informative supervision. The enhanced pruning and initialization is a talented learning process prompting the network to escape from the local minima and re-born from a better start.
   For the second issue, we define a new metric to measure the robustness of gaze estimator, and propose an adversarial training based Disturbance with Ordinal loss (DwO) method to improve it. 
   The experimental results show that our TAT method achieves state-of-the-art performance on GazeCapture dataset, and that our DwO method improves the robustness while keeping comparable accuracy.

\end{abstract}

\section{Introduction}
\begin{figure}[!t]
\begin{center}
\includegraphics[width=0.95\linewidth]{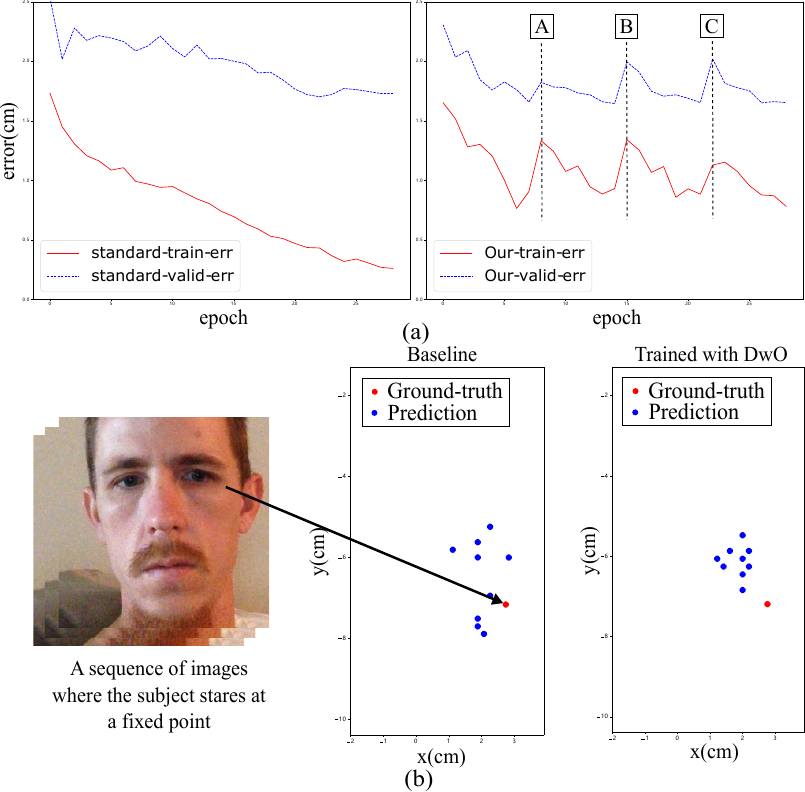}
\end{center}
\caption{(a) A visualization of training (red line) and validation (blue line) error curves. Left: error curves of standard training scheme. The network over-fits on the training set greatly. Right: error curves of our TAT training scheme. Annotations [A-C] denote the re-born points in our iterative TAT. After network re-borns, the error increases and the network converges fast. Our TAT mitigates over-fitting significantly, \ie the gap between two errors is much reduced.
(b) An example of predicted points on phone screen. Left: a series of images where subject stares at a fixed point (red dot). Middle: prediction results (blue dots) trained with standard method. Right: prediction results trained with our method. Using our method, the prediction results have less jitters and are more concentrated comparing to the middle figure. Best viewed in color.}
\label{fig:firstpage_robust}
\end{figure}

Gaze estimation is a task to predict where a person is looking at given the person's full face. The task contains two directions: 3-D~\cite{sugano2017s,zhang2017mpiigaze} gaze vector and 2-D~\cite{krafka2016eye} gaze position estimation. 3-D gaze vector estimation is to predict the gaze vector, which is usually used in the automotive safety.  2-D gaze position estimation is to predict the horizontal and vertical coordinates on a 2-D screen, which allows utilizing gaze point to control a cursor for human-machine interaction. As 2-D gaze technology is often used on smart phone or tablet, the range of head pose and viewing direction are smaller than that of 3-D gaze estimation. In this paper, we focus on 2-D gaze estimation.

Appearance based methods~\cite{sugano2017s,zhang2017mpiigaze,krafka2016eye,fischer2018rt,park2019few} using deep neural network have shown good performance on gaze estimation task.
However, existing methods are still inadequate to be applied in real applications.
On one hand, there is a large performance gap between the training error and the validation error due to severe over-fitting. As shown in Fig. \ref{fig:firstpage_robust} (a)-left, the training error (red curve) is much lower than the validation error (blue curve) using standard training method.
On the other hand, when user stares at one point, the gaze prediction result jitters sharply, as shown in Fig.\ref{fig:firstpage_robust} (b)-middle. The reason is that, given perturbation (though imperceptible by human) on input, CNN have been shown to output notably different result due to its weak robustness~\cite{FGSM,physical,PGD,vulnerable}. It brings great difficulty to gaze based user interaction, \eg the duration time based gaze activation~\cite{magee2015camera} is hard to be realized.

Recent methods~\cite{krafka2016eye,fischer2018rt} capture or generate large dataset to train the network to make the gaze estimator more generalized. However, they still suffer from over-fitting for the following reasons. First, there are still redundant weights in the network whose function is similar with other weights, no matter how large the training set is. Second, there is no training scheme to deal with the useless weights.
As for jitters, although they can be smoothed by temporal smoothing method such as Kalman Filter~\cite{KF}, temporal delay is inevitably introduced, which is intolerable for real-time gaze estimation. There is no existing metric, dataset or method to solve this problem in the training phase.

In this paper, for the first issue, we propose a new Tolerant and Talented (TAT) training scheme to mitigate over-fitting in gaze estimation.
First, we propose a random knowledge distillation framework. It is an iterative tolerant teaching process by randomly selected teachers, which enhances the informative and diverse supervision. It does not cost additional time because all teachers are obtained and employed in one generation.
Second, we propose a cosine similarity pruning method and an aligned orthogonal initialization method. It is a talented learning process which prompts the network to escape from the local minima and to continue optimizing on a better direction. The cosine similarity pruning deletes the useless weights, inheriting the talent of the teacher. The aligned orthogonal initialization method constrains the re-initialized weights orthogonal to each other and aligned to the original distribution, which makes the network re-born from a better start.
Applying these two parts iteratively, we can narrow the performance gap between training set and testing set, achieving higher accuracy in testing set.
For the second issue, we propose a metric to measure the robustness of gaze estimator, and capture a corresponding dataset for experimental validation. Furthermore, in order to improve the robustness, we propose a Disturbance with Ordinal loss (DwO) method which customizes the adversarial training for ordinal loss. To our knowledge, this is the first time to deal with gaze estimation robustness problem.

The proposed method is evaluated on GazeCatpure~\cite{krafka2016eye} and our own dataset. Experimental results show that our TAT training scheme outperforms the state-of-the-art approach on GazeCapture, and that our DwO method reduces jitters on our dataset significantly. The major contributions of this paper can be summarized as follows.
\begin{itemize}
\item A powerful Tolerant and Talented (TAT) training scheme is proposed, which is an iterative random knowledge distillation framework enhanced with cosine similarity pruning and aligned orthogonal initialization. Inheriting the talent from the tolerant teacher, the network can be trained to escape from the local minima, mitigating over-fitting effectively.
\item An adversarial training based Disturbance with Ordinal loss (DwO) method is proposed to address the robustness problem in gaze estimation, which is the first work to our knowledge. Also a quantitative robustness metric for gaze estimation is proposed.
\item Our method achieves state-of-the-art performance on GazeCapture dataset, and enhances significantly the robustness of gaze estimation on our captured dataset with comparable accuracy.
\end{itemize}

\section{Related Work}
\textbf{Gaze Estimation.} In the past few years, gaze estimation draws increasing attention because it provides a great way for human-machine interaction~\cite{yu2019improving,xiong2019mixed,zhang2019evaluation}. Appearance based methods achieve promising results through using deep convolutional neural network (CNN). To make the CNN extract efficient features, Krafka \etal~\cite{krafka2016eye} feed both full face patch and eye patches into the CNN. Sugano \etal~\cite{sugano2017s} use the spatial attention to encode the information about different region of the full face. Most of the existing method use L2 loss to optimize the CNN as gaze estimation is a regression problem~\cite{zhang2017mpiigaze,krafka2016eye}. However, learning from L2 loss suffers from the issue of imbalanced training data~\cite{wang2018hcr}. In this paper, we feed face and eye patches into the network following~\cite{krafka2016eye}, but use ordinal loss~\cite{chen2017using,li2018deep} to predict gaze.

\textbf{Knowledge Distillation.} Teacher-student framework has been demonstrated to improve the performance when the student has the same architecture with the teacher~\cite{yang2019snapshot,yang2018knowledge}. Benefiting from the soft supervision provided by the teacher, student network can learn from the inter-class similarity and potentially lower the risk of over-fitting~\cite{yang2019snapshot}. To reduce the time complexity increased by educating time, Yang \etal~\cite{yang2019snapshot} present snapshot distillation, which enables teacher-student optimization in one generation. However, most of the existing works learn from only one teacher, whose supervision lacks diversity. In this paper, we randomly select a teacher to educate the student.

\textbf{Pruning.} Pruning methods are often used in model compression~\cite{he2017channel,he2018soft}. The main idea is to delete useless weights and keep comparable performance. The weights are pruned according to different criteria such as L1 norm~\cite{li2016pruning}, geometric medium~\cite{he2019filter}. The pruned weights will not be used in the final model. Different from the pruning methods mentioned above, Repr~\cite{prakash2019repr} uses orthogonal coefficient to determine if the weights need to be pruned, and re-initializes the pruned weights orthogonally to continue training the network. However, Repr doesn't consider negative correlation of weights. In addition, the distribution of re-initialized weights is different from that of the pruned weights, which makes the network start from a bad initialization. In this paper, we propose a cosine similarity pruning and aligned orthogonal initialization method to make the network re-born from a better start.

\textbf{Robustness of CNN Models.} While CNN models achieve high accuracy on diverse datasets, they can be easily misled by some small perturbations of their input, which are imperceptible to human, and give wrong predictions. To enhance the robustness, adversarial training are commonly used to train the CNNs, which generates adversarial perturbations according to the gradients of loss w.r.t. the input and adds them to the input as adversarial samples. However, existing methods~\cite{FGSM,physical,PGD,barrage,vulnerable,denoising} mainly focus on classification tasks, yet regression tasks also need to enhance robustness, \eg to reduce prediction jitters. In this paper, we study the robustness of a regression problem, \eg gaze estimation.


\section{Method of Gaze Estimation}
Our method contains two parts. The first part is Tolerant and Talented (TAT) training scheme, which is a knowledge distillation from a randomly selected teachers with a cosine similarity pruning and an aligned orthogonal initialization. The second part is an adversarial training method with the loss of Disturbance with Ordinal (DwO), which generates adversarial samples to enhance the robustness.
Specifically, gaze estimation is a regression problem. L2 Loss is often used to optimize the network's parameters. However, learning from L2 loss suffers from the issue of imbalanced training data~\cite{wang2018hcr}. In this paper, We solve these problem by optimizing ordinal loss, converting regression problem to classification problem. Different from other classification loss such as softmax loss, ordinal loss preserves the property of regression. It provides a larger loss when the prediction is farther away from the ground truth. Ordinal label $y$ is a vector with the length of B, which is converted from a continuous value $gt$, \eg the horizontal or vertical coordinate of the gaze position, following the formula,
\begin{equation}
y^b =
\begin{cases}
1& \text{if $b \cdot BinSize \leq gt$}\\
0& \text{otherwise}
\end{cases}
\label{eq:ordinal_label}
\end{equation}
where $BinSize$ quantifies the prediction range into $B+1$ intervals, $B$ is the bin number, $y^b$ is the b-th component of $y$, indicating whether $b \cdot BinSize$ is smaller than $gt$. The shape of ordinal label is visualized in Fig. \ref{fig:ordinal}

In the training process, we minimize the following formula,
\begin{equation}
\begin{split}
Loss_{ordinal} = \sum_{i=1}^{N}\sum_{b=1}^{B}-(y_i^blog(P_{net}^b(x_i;W))\\
 + (1-y_i^b)log(1-P_{net}^b(x_i;W))),
\end{split}
\label{eq:oridnal_loss}
\end{equation}
where $P_{net}(x_i)$ is the output of the network, \ie a $B$-length vector indicating probability of the corresponding \textit{bin}, $P_{Net}^b(x_i)$ is the b-th component, $x_i$ is a training sample, $W$ are the parameters of the network. $N$ is the image number of the training set.

In the testing process, the sample $x$ is predicted as follows,
\begin{equation}
predict=BinSize\cdot(\sum_{b=1}^{B}\mathcal{I}(P_{net}^b(x;W)\geq0.5) + 0.5),
\label{eq:oridnal_prediction}
\end{equation}
Where $\mathcal{I}(\cdot)$ means indicator function.

\subsection{TAT: Tolerant and Talented Training Scheme}
In this section, we describe the details of the TAT training scheme, as shown in Algorithm \ref{alg:TAT}.

The design idea of the TAT training scheme is to continuously remove the ineffective weights and give the pruned weights another optimization direction, which is non-trivial. \textit{Tolerant} means the opposite of strict, and uses soft label instead of hard label. \textit{Talented} means that the student inherits the talent of the teacher.


The TAT training scheme consists of three parts, a random knowledge distillation (RKD), a cosine similarity pruning (CSP), an aligned orthogonal initialization (AOI). The random knowledge distillation provides smooth supervision from a randomly selected teacher to teach the student network, to prevent over-fitting caused by hard label. The cosine similarity pruning method deletes the ineffective weights, to help the student network escape from the local minima. The aligned orthogonal initialization method re-initializes the pruned weights, to give a better direction to optimize. To enhance the regularization, we also introduce the mixup method to provide an auxiliary loss. The details of the three parts and mixup method are introduced below.
\begin{algorithm}[t]
\caption{TAT Training Scheme.}
\label{alg:TAT}
\begin{algorithmic}[1]
\Require
TeacherList=$\emptyset$, training configuration $\{{\lambda}_{teacher},{\lambda}_{mix},{\lambda}_{hard},p\% \}$, number of epoch $L$, number of mini-generation $K$;
\State Initialize $W_0$;
\For{$k=1,2 \dots K$}
\For{$l=1,2 \dots L$}
\State sample teacher from TeacherList;
\State compute Loss according to Eq. \ref{eq:total loss};
\State update $W_k$;
\EndFor
\State add $W_k$ to TeacherList;
\State compute the $Sim(W_k)$ according to Eq. \ref{eq:cos_similarity};
\State compute top p\% of $Sim(W_k)$, denoted as $\widehat{W_k}$;
\State re-initialize the $\widehat{W_k}$ according to Eq. \ref{eq:merge_BN} \ref{eq:align_weight};
\EndFor
\State \Return $M: y=f(x,W=W_K)$;
\end{algorithmic}
\end{algorithm}

\textbf{Random Knowledge Distillation (RKD).} The function of this module is to provide similarity information in the neighboring gaze positions from teacher's output iteratively. In more detail, the whole generation is split into $K$ mini-generations following~\cite{yang2019snapshot}. Each mini-generation has $L$ epoches except the first one, which contains $L+1$ epoches. The additional $1$ epoch in the first mini-generation is the warmup epoch. When the network is in the training process of $k$, where $k=2 \cdots K$, there are $k-1$ teachers. Similar to the Eq. \ref{eq:oridnal_loss}, the optimization of this part is to minimize the KL divergence between the teacher and the student, following the formula,
\begin{equation}
\begin{split}
Loss_{teacher} = \sum_{i=1}^{N}\sum_{b=1}^{B}-({y'}_i^b\cdot log(P_{net}^b(x_i;{W}_{s}))\\
 + (1-{y'}_i^b)\cdot log(1-P_{net}^b(x_i;{W}_{s}))),
\end{split}
\label{eq:teacher_loss}
\end{equation}
where $y'$ is the output from the teacher's network. ${W}_{s}$ is the parameter of a student network, which needs to be updated.

The teacher in our random knowledge distillation framework provides informative and diverse supervision. The informativeness reflects not only the provided smooth label, but also the quality of the teacher. The diversity reflects the difference of supervision. To keep the informativeness, we remove the teacher whose prediction error is larger than a threshold. To enhance the diversity, we select the teacher randomly.

\textbf{Cosine Similarity Pruning (CSP).} This part is a pruning module, which tries to delete the ineffective filters to help the network escape from the local minima. The metric of the filter selection is the cosine similarity of the filter's weight.
We denote the weight of each layer as $WF$, whose shape is $(N_{out},C,K_w,K_h)$, where $N_{out}$ is the number of the filter in this layer, $C$ is the input channel number, $K_w$ and $K_h$ are the width and height of the filter respectively. The flattened $WF$ is composed of $N_{out}$ vectors with the shape of $C \times K_w \times K_h$. Let $\widetilde{W_{fi}} = W_{fi}/\Vert W_{fi}\Vert$ denote the normalized weights, where $fi=1\cdots N_{out}$. Then we compute the cosine similarity of each filter $fi$ following the formula
\begin{equation}
Sim_{fi}= \frac{(\sum_{col=1}^{N_{out}} ({\widetilde{WF} \times \widetilde{WF}^T - I}))_{row=fi}}{N_{out}},
\label{eq:cos_similarity}
\end{equation}
where $\widetilde{WF} \times \widetilde{WF}^T$ is a matrix of size $N_{out} \times N_{out}$. $Sim_{fi}$ means summing all the cosine similarities between filter $fi$ and all the other filters in this layer. Following~\cite{prakash2019repr}, we compute the metric in a single layer, and the ranking is computed over all the filters in the network. However, there are two different points from the method in~\cite{prakash2019repr}. First, we don't use the absolute value. In fact, the negative correlation, i.e. cosine similarity is equal to $-1$, should be considered differently from the positive correlation, because the ReLU layer following the convolution layer suppresses the negative values. Second, we add a constraint that each layer's pruning ratio can't be larger than a threshold $p_{max}\%$. Otherwise, most of the filters in one layer may be pruned, and the training starts from scratch. We prune the weights with top $p\%$ $Sim_{fi}$.

\textbf{Aligned Orthogonal Initialization (AOI).} This part gives a new initialization of the pruned weights, to help the network continue optimizing on a better direction. There are two principles to design the re-initialization method. First, the re-initialized weights need to have low cosine similarity according to the formula \ref{eq:cos_similarity}. It suppresses repeated pruning in the next mini-generation. Second, the L2 norm of re-initialized weights need to match with that of the pruned ones. Specifically, if the L2 norm of re-initialized weights is very small, there is little contribution coming from the re-initialized weights. On the opposite, if the L2 norm is very large, the network is likely to collapse. Thus, we re-initialize the pruned weights in an orthogonal way with three steps. First, we get $W\_Re$, whose shape is $(N_{pruned},C,K_w,K_h)$, by applying QR decomposion\footnote{https://pytorch.org/docs/stable/nn.html\#torch-nn-init} on $WF$. Second, we compute the weights adjusted with BN parameters, according to the formula,
\begin{equation}
W\_adj_{fi}= \frac{W_{fi}\cdot BN\_scale_{fi}}{\sqrt{BN\_var_{fi}}},
\label{eq:merge_BN}
\end{equation}
where $BN\_scale$ and $BN\_var$ are the $scale$ and $variance$ in the batch normalization layer, $fi$ means the fi-th pruned filter.
Third, the $W\_Re$ are aligned following the formula,
\begin{equation}
W\_aligned_{fi}= \frac{W\_Re_{fi}}{\Vert W\_Re_{fi}{\Vert}_2} \cdot Scalar_{aligned}.
\label{eq:align_weight}
\end{equation}
The orthogonality can be preserved if all the vectors in $W\_Re$ are multiplied by a scalar. $Scalar_{aligned}$ is sampled from a $(min(\{\Vert W\_adj_{fi}{\Vert}_2,\forall fi\}),max(\{\Vert W\_adj_{fi}{\Vert}_2, \forall fi\}))$ uniform distribution.

\textbf{Mixup.} Mixup~\cite{zhang2017mixup} method generates virtual samples following the formula,
\begin{equation}
\begin{split}
x_{mix}= \alpha \cdot x_{i} + (1 - \alpha) \cdot x_{j}, \\
label_{mix} = \alpha \cdot label_{i} + (1-\alpha)\cdot label_{j},
\end{split}
\label{eq:mixup}
\end{equation}
where parameter $\alpha$ is sampled from beta distribution in~\cite{zhang2017mixup}, $x_i$ and $x_j$ are the data in the training set. Mixing the feature in the last layer doesn't work because the mixed samples are treated as vicinity samples~\cite{zhang2017mixup}. However, in an ordinal ranking task, mixing the feature in the last layer is treated as a regularization term because of the regression property. We sample $\alpha$ from a (0, 1) uniform distribution. We will show that it improves the performance in the Section \ref{Ablation study}.

Finally, the total loss is
\begin{equation}
\begin{split}
Loss_{total} = {\lambda}_{hard} \cdot Loss_{hard} + {\lambda}_{mix} \cdot Loss_{mix} \\
+ {\lambda}_{teacher} \cdot Loss_{teacher},
\end{split}
\label{eq:total loss}
\end{equation}
where $Loss_{hard}$ and $Loss_{mix}$ are ordinal losses whose ground-truth comes from the original data and mixed data, respectively.


\subsection{DwO: Disturbance with Ordinal Loss}\label{DwO}
In this section, we describe the details about improving the robustness of gaze estimators.
Firstly, we define an evaluation metric to measure the robustness. Secondly, we capture a dataset, named GazeStare, for evaluating this metric. Finally, we propose an adversarial training based Disturbance with Ordinal loss (DwO) method to improve the robustness of gaze models.

The idea of this method stems from the key observation of notable jittering of gaze estimation results as shown in Fig. \ref{fig:firstpage_robust}-(b). While there is no existing paper studying the gaze estimation robustness of jittering, it has tremendous influence on practical use cases. Thus, this paper propose an evaluation metric, a dataset and a training method regarding to the robustness of gaze estimation.

\textbf{MSD evaluation metric.} In general, the papers studying the adversarial robustness are based on classification tasks and they use classification accuracy as metric~\cite{FGSM,physical,PGD,barrage,vulnerable,denoising}. However, gaze estimation is a regression problem and we aim to measure the robustness of gaze estimation results when a person stares at a fixed point. To this end, we propose Mean Standard Deviation (MSD) as an evaluation metric, which is defined as
\begin{equation}
\mu(S_j) = \frac{1}{M}\sum_{i = 1} ^Mg(x_i)  \label{eq_MSD1}
\end{equation}
\begin{equation}
\sigma(S_j) = \sqrt{\frac{1}{M}\sum_{i = 1} ^M \Vert g(x_i) - \mu(S_j) {\Vert}_2^2 } \label{eq_MSD2}
\end{equation}
\begin{equation}
MSD(\mathcal{D}) = \frac{1}{N_s}\sum_{j = 1} ^{N_s} \sigma(S_j) \label{eq_MSD3}
\end{equation}
with $S_j = \{x_1,x_2,...,x_M\}$ being a set of consecutive images where a person stares at a fixed point. We refer to $S_j$ as a \textit{sequence} hereafter. $g(\cdot)$ means the predicted gaze coordinates. $N_s$ denotes the number of sequences in dataset $\mathcal{D}$. The MSD calculates the average standard deviation of prediction result of all the sequences in the dataset.

\textbf{GazeStare dataset.} Currently no existing gaze dataset contains large number of sequences with many images where subjects stare at fixed points continuously, yet in practical use this kind of sequences are easily acquired. For instance, 47\% of sequences in GazeCapture dataset have less than 5 images each. Evaluating MSD with such datasets is not statistically convincing. Thus, we collect a dataset called GazeStare containing sequences with abundant images. GazeStare consists of images from 14 subjects. A subject stares at 8-10 points and about 210 frames are captured for each point. To have less noise data, eye-blinking frames are removed from the dataset. As a result, GazeStare contains 26427 images.

\begin{figure}[t]
\begin{center}
\includegraphics[width=1.0\linewidth]{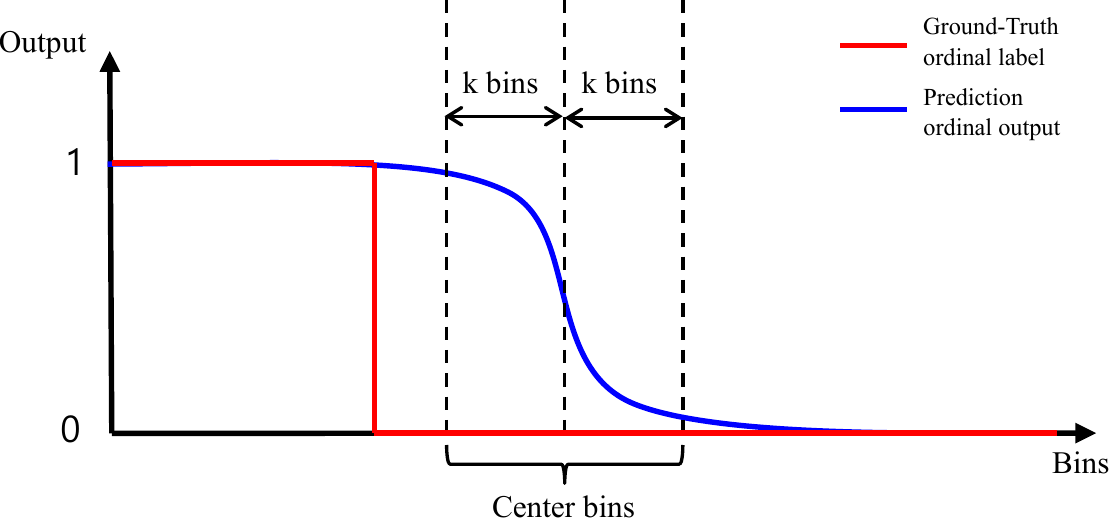}
\end{center}
\caption{Illustration of our ameliorated usage of ordinal loss when calculating the gradients. Instead of using all the bins of the ordinal output, we only leverage \textit{center bins}, \ie the $2k$ bins neighboring to the prediction. Best viewed in color.}
\label{fig:ordinal}
\end{figure}

\textbf{Training method.} To achieve better robustness of gaze estimator, we propose an adversarial training based method customized for the ordinal loss. In training phase, we generate adversarial perturbation based on the Projected Gradient Descendant (PGD)~\cite{PGD} method and add them to original inputs to imitate small pixel-wise variation, \eg illumination variation. The PGD is a strong iterative method to generate perturbation, which can be written as

\begin{equation}
x_{0}^{adv} = x \label{eq_MSD4}
\end{equation}
\begin{equation}
x_{t+1}^{adv} = clip(x_t^{adv} + \gamma \cdot sign(\nabla_{x_t^{adv}}L(P_{net}(x_t^{adv}),y)), \epsilon) , \label{eq_MSD5}
\end{equation}
where $\gamma$ is the step length, and $t$ denotes the iteration step, which reaches $T$ to finish the perturbation generation. $\epsilon$ is $L_{\infty}$ norm constraint for the adversarial samples. $L(\cdot)$ is the loss function \ie ordinal loss in our gaze estimator. $P_{net}(\cdot)$ is the output of our network.

To make the ordinal loss harmonize with the adversarial training, we ameliorate the usage of ordinal loss. More specifically, as illustrated in Fig. \ref{fig:ordinal}, when calculating the gradients of ordinal loss w.r.t input, we only leverage $2k$ bins near the prediction of the ordinal output to make the gradients more reasonable (we refer to these bins as \textit{center bins} hereafter). The reason is that, when the prediction varies slightly, it's mainly the excitation of the bins near the prediction that varies, so we don't take bins far from the prediction into account when calculating the gradients. This design makes the adversarial samples more meaningful and therefore results in better performance, which is shown in the ablation study.

In the training process, we mix original samples and adversarial samples in order to maintain the estimation accuracy. The ratio of original samples in the training set is denoted as Org.\%.

\section{Experiment}

In this section, we first introduce the datasets on which the method is evaluated, and then detail the settings of our method. After that, we compare with two methods. The first one is iTracker~\cite{krafka2016eye}, which is the state-of-the-art in 2-D gaze estimation, and the second one is SD~\cite{yang2019snapshot}, which is the state-of-the-art general knowledge distillation method. We reproduce their method for gaze estimation task. Ablation study is also conducted to show the contribution of each part of our work.

\subsection{Dataset and Configuration}
\textbf{GazeCapture dataset.} GazeCapture dataset~\cite{krafka2016eye} is a 2-D gaze dataset captured with iphone and ipad in different orientations, which contains 1,490,959 valid frames (both eyes and faces detected) from 1471 subjects. The dataset is divided into train, validation, and test splits consisting of 1271, 50, and 150 subjects, respectively.

\textbf{GazeCN dataset.} In Section \ref{DwO}, we introduce our GazeStare dataset designed for measuring robustness of gaze estimator. To keep consistent with the distribution of that dataset for better validating the effectiveness of our DwO method, we also capture the GazeCN dataset. GazeCN is a 2-D gaze dataset which consists of images from 290 subjects. The images are collected with Galaxy S8+, under different illumination conditions, at different body postures and head poses. We split the subjects into two non-overlap parts, which contain 263 and 27 subjects respectively, as train set and test set. As a result, the train set and test set have 262,400 and 43,380 images, respectively.

\textbf{Network Configuration.} Following~\cite{krafka2016eye}, the input of the network contains $3$ parts with the size of $64\times64$, \ie face patches, left and right eye patches. We use the architecture in~\cite{yi2014learning} instead of AlexNet~\cite{krizhevsky2012imagenet} to enhance the fitting ability of the network. The output of each sub-network is a vector of $128$ dimension. After these three 128-D vectors are combined, a fully connection layer is followed whose size is $384 \times 128$. The dimension of final feature is $128$. For GazeCapture dataset~\cite{krafka2016eye}, we set bin number $B$ to $133$ for both horizontal and vertical coordinates. For our GazeCN dataset, we use the $72$ and $98$ bin number for horizontal and vertical coordinates respectively.

There are some hyper-parameters in the training scheme. We set ${\lambda}_{teach}$, ${\lambda}_{mix}$, ${\lambda}_{hard}$ to 0.6, 0.4, 0.2 respectively. In random knowledge distillation, we use $1$ epoch to warmup, $7$ epoches for knowledge distillation. The number of mini-generation $K$ is $5$. In cosine similarity pruning, we set the pruning ratio $p\%$ to $20\%$, the max ratio for each layer $p_{max}\%$ to $50\%$. In DwO, we set the $\epsilon$ to 3, the $\gamma$ to 1.

\subsection{Comparison with state of the art}
We first show the results on GazeCapture dataset~\cite{krafka2016eye}. As shown in Tab. \ref{tab:GazeCapture_SOTA}, compared with iTracker~\cite{krafka2016eye}, our result outperforms in both iphone dataset and ipad dataset, achieving 4.8\% and 5.3\% error reduction respectively. Compared with SD~\cite{yang2019snapshot} which is reproduced by us for gaze estimation, our error is a little higher in ipad set, but lower in iphone set and the total set.
\begin{table}[ht]
\centering
\begin{tabular}{|c|c|c|c|}
\hline
         & iphone & ipad & total\\
\hline
iTracker~\cite{krafka2016eye}     & 1.86 & 2.81 &2.05\\
\hline
SD~\cite{yang2019snapshot}   & 1.81 & 2.61 & 1.97\\
\hline
TAT    & 1.77 & 2.66 & 1.95\\
\hline
\end{tabular}
\caption{Test errors (in cm.) on GazeCapture dataset. First row is the result reported in~\cite{krafka2016eye}. The second row is the result reproduced by us using the method in~\cite{yang2019snapshot}. The last row is the result of our TAT training scheme.}
\label{tab:GazeCapture_SOTA}
\end{table}

Note that the image resolution we used are $64\times64$ for all three patches, since we are interested in the performance gain from the training algorithm.
Larger image size, deeper network and model ensemble are very effective for boosting the model accuracy.
For example, Kannan~\cite{kannan2017eye} uses the image size of $448\times448$ to achieve the error of $1.75cm$ on GazeCapture dataset.
These techniques can be integrated with our TAT method to further improve the gaze estimation accuracy.

\subsection{Ablation study}\label{Ablation study}
In ablation study, we use the subset of GazeCapture named iphone orientation $1$, which contains about 400k training frames, 19k validation frames and 55k testing frames.

%
%

\textbf{Analysis of Random Knowledge Distillation (RKD).}
In Tab. \ref{tab:ablation_KD}, we fix the pruning method and re-initialization method as our proposed CSP and AOI. The result of ``Finetune'' means that we prune the useless weights and re-initialize them, then fine-tune the network using hard loss and mix loss only in Eq. \ref{eq:total loss}. We can see that the error reduction from the MG.0 to MG.1 is lower than other methods. With the increasing of the mini-generation step, the error doesn't decrease because it lacks teacher's supervision. The error of ``LastOne'' stops decreasing after the MG.3. We guess that the teacher's quality, \ie the performance of teacher, becomes low with the increasing of the mini-generation step. As we use the average output of the teachers, which can improve the teacher's performance, we get more error reduction as shown in the result of ``Our mean''. ``Our Best'' provides the best teacher, but the result is worse than "Our random", which demonstrates that the teacher's diversity is also important.

\begin{table}[ht]
\centering
\begin{tabular}{|c|c|c|c|c|c|}
\hline
                 & MG.0 & MG.1 & MG.2 & MG.3 & MG.4\\
\hline
Finetune        &1.81 &1.80 &1.78 &1.79 &1.79 \\
\hline
LastOne    &1.80 &1.77 &1.76 &1.75 &1.77\\
\hline
Our mean         &1.80 &1.77 &1.76 &1.75 &1.75\\
\hline
Our best         &1.79 &1.77 &1.77 &1.76 &1.77\\
\hline
Our random         &\textbf{1.79} &\textbf{1.77} &\textbf{1.75} &\textbf{1.74} &\textbf{1.73}\\
\hline
\end{tabular}
\caption{The test errors (in cm.) of different knowledge distillation methods. ``Finetune'' means that there is no teacher, and the supervision comes from the hard label. ``LastOne'' means that we use the teacher in the last mini-generation, which is similar to~\cite{yang2019snapshot}. ``Our mean'' means that the supervision are the mean output of all the model in the previous mini-generations. ``Our best'' uses the best one \ie lowest error, in the previous mini-generations. ``Our random'' means that we randomly select a model as the teacher from the previous mini-generations. ``MG.\#'' means the mini-generation step.}
\label{tab:ablation_KD}
\end{table}

\textbf{Analysis of Cosine Similarity Pruning (CSP).}
In Tab. \ref{tab:ablation_pruning}, we fix RKD and AOI, and compare different pruning methods. The result of ``Repr'' shows the error is not stable. There are two reasons. First, ``Repr'' computes the absolute value of cosine similarity, which causes that many weights near the input layer are pruned, because the weights near the input layer have a large negative correlation as demonstrated in~\cite{shang2016understanding}. Second, if we don't constrain the pruning ratio in each layer, most of the weights in the same layer are pruned leading to a large influence for the next training. This instability is also demonstrated in the result of ``Scratch'', which doesn't use pruning method and re-initializes all the weights in the student network. It means that the student doesn't inherit anything from the teacher. In our CSP, we consider that weights with large negative correlation are also important. In addition, we constrain the pruning ratio of each layer to be lower than a threshold \ie $p_{max}\%$ in order to make the student network inherit talent from the teacher. Result of our method shows that the error is reduced step by step.

\begin{table}[ht]
\centering
\begin{tabular}{|c|c|c|c|c|c|}
\hline
                 & MG.0 & MG.1 & MG.2 & MG.3 & MG.4\\
\hline
Repr   &1.80 &1.78 &1.77 &1.78 &1.77 \\
\hline
Scratch  &1.79 &1.82 &1.80 &1.80 &1.80\\
\hline
Our CSP  &\textbf{1.79} &\textbf{1.77} &\textbf{1.75} &\textbf{1.74} &\textbf{1.73}\\
\hline
\end{tabular}
\caption{The test errors (in cm.) of different pruning methods. ``Repr'' means that we use the pruning method in ~\cite{prakash2019repr}. ``Scratch'' means that we re-initialize all the weights. ``Our CSP'' means that we use our CSP method.}
\label{tab:ablation_pruning}
\end{table}

\textbf{Analysis of Aligned Orthogonal Initialization (AOI).}
\begin{figure}[t]
\centering
\includegraphics[width = 1.0\linewidth]{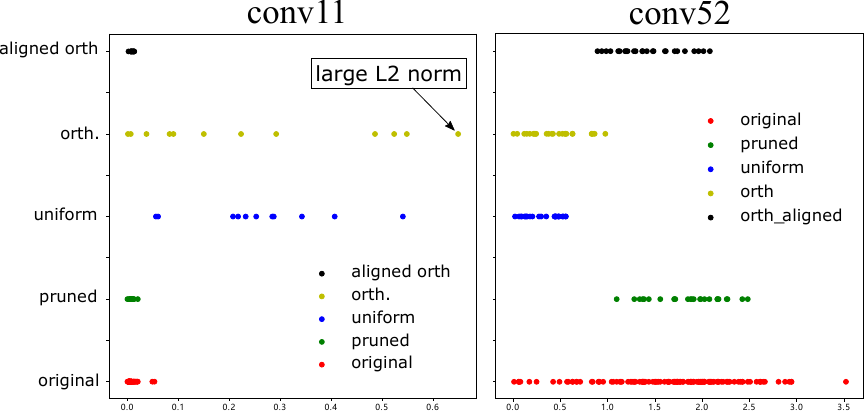}
\caption{Visualization of the distribution of weights re-initialized by different methods. The x-axis means the value of L2 Norm. We visualize the range of L2 Norm of different methods in different positions of y-axis. The left figure is the result of the layer ``Conv11''. The right figure is the result of the layer ``Conv52''. Best viewed in color.}
\label{fig:reinit_fig}
\end{figure}
In Tab. \ref{tab:ablation_reinit}, we compare different re-initialization methods, fixing RKD and CSP. It shows that the``Orth.'' method doesn't converge after MG.1. The result of uniform re-initialization is almost unchanged after MG.1. To explain the reason of these phenomenons, we visualize the L2 Norm of the weights, as shown in Fig. \ref{fig:reinit_fig}. The figure shows the L2 Norm range of the weights. In layer ``Conv11'', which is the first layer of the network, the L2 Norm of both the pruned weights and original weights are lower than $0.1$, which are much smaller than both of the weights re-initialized by ``Uniform'' and ``Orth''. Especially, there is one filter re-initialized by ``Orth.'' whose weights have much larger L2 Norm than the pruned weights. It causes an abnormal output of layer ``Conv11'', which influence all successive layers, leading to the network collapse and divergence. In the layer ``Conv52'', the L2 Norm of the pruned weights has a higher bound than both of the weights re-initialized by ``Uniform'' and ``Orth.''. It means that the re-initialized weights make little contribution to the network. Though the weights with smaller L2 Norm can be updated during training, they still have a worse starting point compared with ``Our AOI'' method. On the contrary, the weights re-initialized by ``Our AOI'' have the same bound of L2 Norm with the pruned ones. 
\begin{table}[ht]
\centering
\begin{tabular}{|c|c|c|c|c|c|}
\hline
                 & MG.0 & MG.1 & MG.2 & MG.3 & MG.4\\
\hline
Orth.  &1.79 &N/A &N/A &N/A &N/A\\
\hline
Uniform    &1.81 &1.77 &1.78 &1.77 &1.77 \\
\hline
Our AOI  &\textbf{1.79} &\textbf{1.77} &\textbf{1.75} &\textbf{1.74} &\textbf{1.73}\\
\hline
\end{tabular}
\caption{The test errors (in cm.) of different re-initialization methods. ``Orth.'' and ``Uniform'' mean the orthogonal and uniform re-initialization respectively. ``Our AOI'' means our proposed aligned orthogonal initialization method. }
\label{tab:ablation_reinit}
\end{table}

\textbf{Analysis of DwO.} In Tab. \ref{tab:table_robust_1}, we report the performance of the model trained with DwO on GazeCN dataset. The error is tested on GazeCN test set. The MSD is tested on the proposed GazeStare dataset to measure the robustness. We fix $\gamma=1$ in all the setups. As we diminish the number of center bin in ordinal loss, the model with 8 center bins achieves the best performance (low error and low MSD), which validates the effectiveness of the center bin loss's design. The reason that this specific number works best is that, for a majority of samples, bins with output between 0.1 and 0.9 are in range of 8 bins around the prediction, so generating adversarial perturbation on these bins is the most reasonable. On bottom half of Tab. \ref{tab:table_robust_1}, as we increase the PGD iteration number (T) and decrease the original data\%, we sacrifice more accuracy to get better robustness. Here we recommend the setup with T=1 and Org.\%=90\%, which achieves 19.2\% drop of MSD with comparable accuracy to the baseline, which is trained with a standard training scheme. Using the model trained with our DwO method, the user can experience significant reduction of jitters and barely feel the drop on accuracy.
\begin{table}[t]
\begin{center}
\begin{tabular}{l|ccc||cc}
\hline
\multirow{2}{*}{Model} & \# Ctr. & \multirow{2}{*}{T} & \multirow{2}{*}{Org.\%} & Err. & MSD  \\

&\ bin&&&(cm.)& (cm.) \\
\hline \hline
Baseline&&&&1.15&0.42\\
\hline \hline

\multirow{8}{*}{DwO}&All&1&90&1.20 &0.36 \\
 &32&1&90&1.22 &0.35\\
 &16&1&90&1.20 &0.34 \\
 &8&1&90&\textbf{1.18}&\textbf{0.34}\\
\cline{2-6}

 &8&2&90&1.25 &0.30 \\
 &8&3&90&1.28 &0.29 \\
\cline{2-6}

 &8&1&80&1.22 &0.30 \\
 &8&1&50&1.30 &0.29 \\
 &8&1&0&1.44 &0.27 \\
\hline

\end{tabular}
\end{center}
\caption{GazeCN and GazeStare: Performance of the network trained with DwO with different setups. \#Ctr. bin refers to the number of center bins. T refers to the PGD iteration number. Org.\% stands for the percentage of the original training data. Error and MSD are the lower the better. The boldface line shows that our DwO method improves the robustness significantly while keeping comparable accuracy with the baseline. }
\label{tab:table_robust_1}
\end{table}


\section{Conclusion}
In this paper, we claim that to achieve accurate gaze estimation in the real work, there are still two problems, \ie performance gap between training set and testing set, and robustness. To solve the over-fitting problem, we propose a Tolerant and Talented (TAT) training scheme. The TAT training scheme is a tolerant teaching and talented learning method. The tolerant teaching process is based on a knowledge distillation framework, which providing diverse and informative supervision. The talented learning process benefits from cosine similarity pruning and aligned orthogonal initialization, which prompts the network to escape from the local minima and to continue optimizing on a better direction.
For the robustness problem, we propose a metric and a dataset to measure the robustness of gaze estimator. Further, we propose a Disturbance with Ordinal loss (DwO) method to improve the robustness by customizing adversarial training for ordinal loss.
We demonstrate that our TAT method achieves state-of-the-art on GazeCapture dataset, and that our DwO method improves gaze estimator's robustness significantly while keeping comparable accuracy. In the future, we also plan to employ our methods on other regression tasks.

{\small
\bibliographystyle{ieee}
\bibliography{egbib}
}

\end{document}